\documentclass[10pt, a4paper]{article}
\usepackage{lrec2022} 
\usepackage{multibib}
\usepackage{graphicx}
\usepackage{tabularx}
\usepackage{soul}
\usepackage{svg}
\usepackage{fancyvrb}
\usepackage{comment}
\usepackage{titlesec}
\titleformat{\section}{\normalfont\large\bfseries\center}{\thesection.}{1em}{}
\titleformat{\subsection}{\normalfont\SmallTitleFont\bfseries\raggedright}{\thesubsection.}{1em}{}
\titleformat{\subsubsection}{\normalfont\normalsize\bfseries\raggedright}{\thesubsubsection.}{1em}{}
\renewcommand\thesection{\arabic{section}}
\renewcommand\thesubsection{\thesection.\arabic{subsection}}
\renewcommand\thesubsubsection{\thesubsection.\arabic{subsubsection}}

\usepackage{epstopdf}
\usepackage[utf8]{inputenc}

\usepackage{hyperref}
\usepackage{xstring}

\usepackage{color}
\usepackage{amssymb}

\usepackage{caption}
\usepackage{subcaption}
\usepackage{nicematrix}
\usepackage{booktabs}

\newcommand{\slurk}{\textit{\texttt{slurk}}}
\newcommand{\Slurk}{\textit{\texttt{slurk}}}
\newcommand{\openvidu}{\textit{OpenVidu}}
\newcommand{\amt}{\textsc{amt}}
\newcommand{\parlai}{parl\textsc{AI}}
\newcommand{\html}{\textsc{html}}
\newcommand{\js}{\textsc{JavaScript}}

\title{The \slurk \ Interaction Server Framework:\\Better Data for Better Dialog Models}

\name{Jana Götze, Maike Paetzel-Prüsmann, Wencke Liermann,\\{\bf \large Tim Diekmann}, {\bf \large David Schlangen}}

\address{Foundations of Computational Linguistics, Department Linguistics \\
         University of Potsdam, Germany \\
         \{firstname.lastname\}@uni-potsdam.de, diekmann.tim@gmail.com\\}

\abstract{
This paper presents the \slurk \ software, a lightweight interaction server for setting up dialog data collections and running experiments. \Slurk \ enables a multitude of settings including text-based, speech and video interaction between two or more humans or humans and bots, and a multimodal display area for presenting shared or private interactive context. The software is implemented in Python with an \html \ and \js \ frontend that can easily be adapted to individual needs. It also provides a setup for pairing participants on common crowdworking platforms such as Amazon Mechanical Turk and some example bot scripts for common interaction scenarios.
 \\ \newline  }

\begin{document}

\maketitleabstract

\section{Introduction}

Much of NLP's breakthroughs in recent years is based on data-driven learning methods. Data-hungry machine learning algorithms to model Visual Question Answering \cite{das2017} or Vision and Language Navigation Tasks \cite{krantz2020} are fed with crowdsourced data, which is fast and affordable to obtain, as crowdworkers do not have to be brought to the lab for common tasks like labeling, captioning images or producing navigation instructions. Data for the subfield of dialog modelling requires at least two crowdworkers to be involved, something that is a non-standard use case for the most popular crowdsourcing platforms as it requires coordinating two or more workers to find a common timeslot to work on a task.

The \slurk \ tool adds to a body of frameworks and tools \cite{healey2003,manuvinakurike2015,miller2017,schlangen2018a} that facilitate data collection for training and testing dialog models where it is often necessary for researchers to set up their own data collection that satisfies their specific needs, e.g., to cover some specific domain or dialog phenomenon. \Slurk \ allows to set up experiments for human-human or human-machine interaction, with no limitation on the number of participants for a given setup. Possible interaction channels include text as well as audio and video. Dialog games – consisting of an interaction setting such as a text or audio channel, a certain context to refer to, and a task to solve – can be created to include multimodal context such as images or interactive tools in which participants can manipulate the context together.

In this paper, we describe the \slurk \ software, a modular tool for collecting multimodal dialog data that is integratable with crowdworking platforms to pair up participants on demand. We explain how \slurk \ can be set up to constrain the interaction channel in a number of ways as well as to manipulate the visual context for each participant. Section~\ref{sec:purpose} details the purpose of the software, Section~\ref{sec:related} describes related frameworks, and Section~\ref{sec:architecture} introduces the \slurk \ architecture and system features in detail. In Section~\ref{sec:example_settings} we demonstrate how these features can be used to set up data collections and experiments, in the lab or via crowdsourcing.
\section{Goals}
\label{sec:purpose}

\begin{figure*}
    \begin{center}
    \includegraphics[width=0.9\textwidth,trim=70 110 88 50,clip]{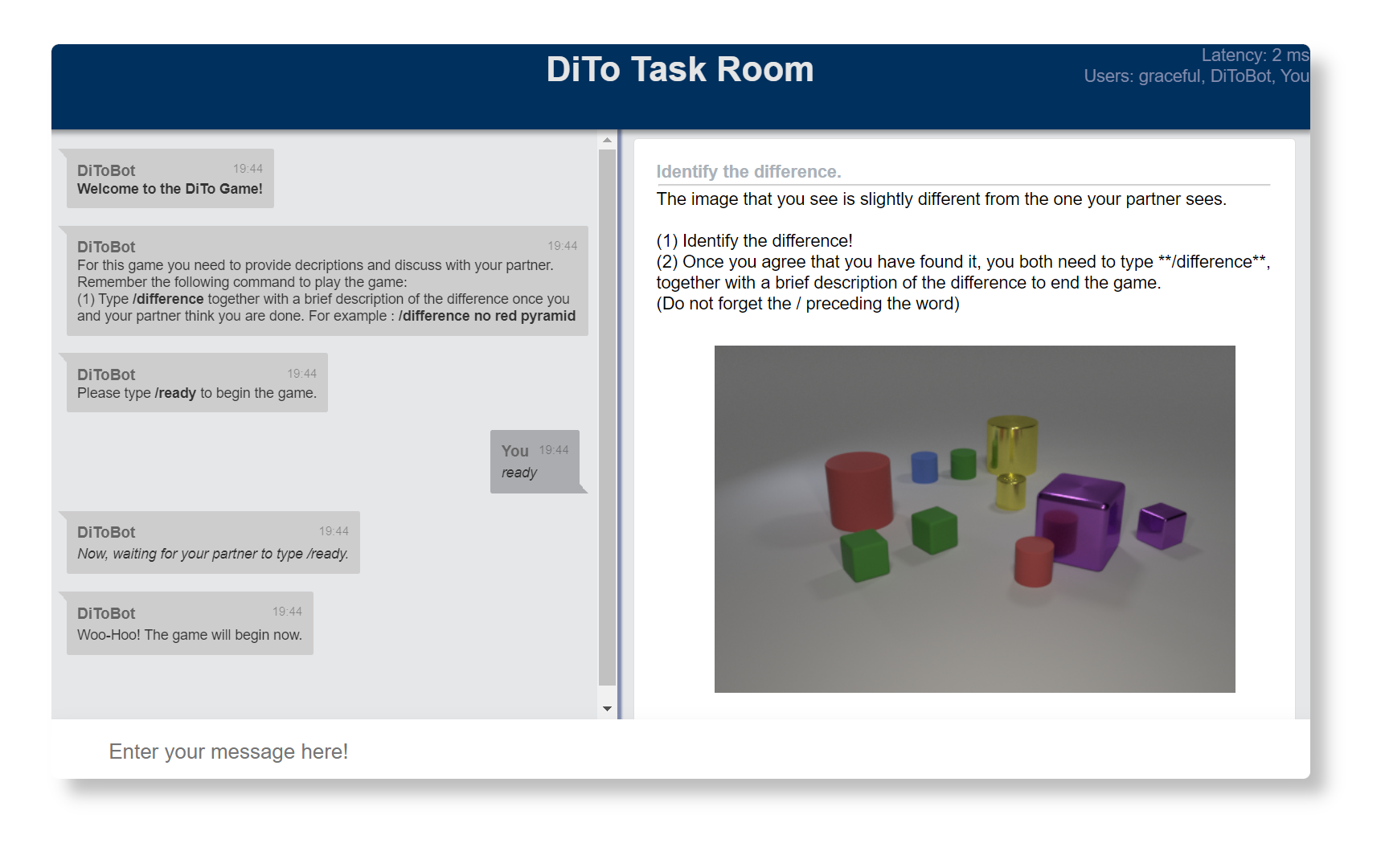}
    \end{center}
    \caption{One participant's browser view in the DiTo task. The browser view is separated into the {\em chat area} on the left and the {\em display area} on the right. The other participant in this task sees a different image. The top right corner shows who is present in the room. Images are taken from the \textsc{clevr} dataset \protect\cite{johnson2017}.}
    \label{fig:dito_image}
\end{figure*}

The main purpose of \slurk \ is to provide a framework that is flexible and modular to set up a variety of dialog tasks in order to both collect data from human conversations as well as test existing dialog models with human evaluators. Dialog context, such as images or interactive buttons, as well as the interaction channel can be manipulated in a number of ways as we outline in this paper. Figure~\ref{fig:dito_image} shows an example interface for a dialog task, with the chat area on the left, showing the dialog history, and what we call the display area on the right, providing the visual context.

Developing new dialog models for different settings often requires researchers to first collect data of humans performing both sides of the task in order to fully understand the parameters of what they are modelling, e.g., what does the conversation look like when participants have different roles, when they do not share the visual context or when the task formulation changes slightly? Then, once a model has been developed, it is imperative to test it with human users. While many metrics exist that evaluate dialog along a number of dimensions, interactions with and judgments from humans are vital to understand a model's scope and detect its limitations.

\Slurk \ aims to be useful for both understanding human-human conversations as well as evaluating dialog models and is designed to be extended for new settings. For example, we can use the display area to connect users to external tools and send information via the \slurk \ server, thus synchronizing all information and logs in one place.
\section{Related work}
\label{sec:related}

The three platforms that are most similar to \slurk \ are the \parlai \ tool \cite{miller2017}, the DiET toolkit \cite{healey2003} and the Pair Me Up Web Framework \cite{manuvinakurike2015}.

The \parlai \ platform \cite{miller2017} was built with the goal of supporting the creation of general open-domain chatbots in mind. It is a text-based dialog platform that lets developers of language models test their trained models on many different tasks to which the platform provides seamless access. The framework can also be used for training models and sharing datasets. \parlai \ has been used mainly for open-ended dialog, e.g., to study what factors make for a good conversation \cite{see2019}.

The Dialogue Experimentation Toolkit (DiET) \cite{healey2003} focusses on studying human-human text-based dialog. The toolkit aims at studying manipulations to the interaction settings, e.g., by deleting, changing, or adding turns. It also provides a {\sc gui} interface for defining such interventions. Some DiET features can be run via the open-source chat messenger Telegram, making it easily accessible for non-lab participants. The toolkit has been used for example to study the role of laughter in chats by inserting artificial laughter tokens into turns \cite{maraev2020}.

The Pair Me Up Web Framework (PMU) \cite{manuvinakurike2015} has originally been developed for collecting human-human spoken conversations over the web and was later extended to allow for autonomous bots to be paired with human conversation partners as well. Although it has so far been used only in one particular interactive setting (RDG-Image \cite{paetzel2014a}) and is not actively maintained anymore, the general technology developed for pairing participants as well as recording their audio and synchronizing it with events in the game interface could be applied to other domains.

While the first two platforms focus on two very different goals – building an open-ended chatbot vs.\ studying human-human interactions – they share some features. \parlai\ and DieT are text-based and comprise a display area to present visual context to the human dialog partners in the form of images. Image displays are configurable in both, but in contrast to \slurk, no interactive elements such as buttons can be part of the interaction context. The third platform, PMU, like \slurk, allows for an interactive visual context, the pairing of both human and artificial conversation partners as well as the potential for spoken conversations; it has however not been developed into a general purpose tool (and is not in active development), and lacks the flexibility that \slurk\ offers.

\parlai \ dialogs follow a strict turn-taking regime in which participant and chatbot turns strictly alternate. Turns are displayed in full as the focus is on the generated language. DiET is more flexible with respect to turn-taking. It allows free turn-taking between participants and implements, in addition to sending messages turn by turn, a \textsc{wysiwyg} mode that simulates incrementality by sending typed characters to the other participant immediately, and lets them fade after some time. \Slurk \ implements a similar mode called \textit{live-typing} that can send messages character by character but does not let them fade from the display area. PMU allows for communication via spoken language only and does not enable experimental restrictions on turn-taking.
\Slurk \ also implements a number of additional options to control the interaction setting, such as incorporating mouse clicks or annotation elements and an audio and video channel that can be restricted, e.g., by enforcing a push-to-talk turn-taking setting.

For setting up data collections or experiments, \parlai \ provides an \amt \ interface, and DiET an interface to the Telegram chat messenger. \Slurk \ and PMU provide a pairing setting that can be used with either a crowdsourcing platform like \amt \ or via a desktop browser directly. Table~\ref{tab:platforms} summarizes some of the four tools' features, providing an overview of the main differences between them.

\begin{table}
  \setlength{\tabcolsep}{3.5pt}
  \begin{tabular}{|l|c|c|c|c|}
  \hline
   & {\bf\parlai} & {\bf DiET} & {\bf PMU} & {\bf\slurk}\\\hline
  \textbf{Interaction} &&&&\\
  ~via text & $\checkmark$ & $\checkmark$ & & $\checkmark$\\
  ~via audio &  & & $\checkmark$ & $\checkmark$ \\
  ~via video &  &&& $\checkmark$ \\
  ~flex.\ turn-taking && $\checkmark$ & $\checkmark$ & $\checkmark$ \\
  ~human-human && $\checkmark$ & $\checkmark$ & $\checkmark$ \\
  \hline
  \textbf{Dialog context} &&&&\\
  ~images & $\checkmark$ & $\checkmark$ & $\checkmark$ & $\checkmark$ \\
  ~config.\ layout & $\checkmark$ && ($\checkmark$)& $\checkmark$ \\
  ~interact.\ elements &&& $\checkmark$& $\checkmark$ \\
  \hline
  \textbf{Data collection} &&&&\\
  ~\amt \ integration & $\checkmark$ && $\checkmark$& $\checkmark$\\
  ~Telegram integr. && $\checkmark$ &&\\
  ~Models included\ & $\checkmark$ &&&\\
  \hline
  \end{tabular}
  \caption{Features of the dialog platforms \parlai \ \protect\cite{miller2017}, DiET \protect\cite{healey2003}, Pair Me Up \protect\cite{manuvinakurike2015} and \slurk.}
  \label{tab:platforms}
\end{table}

Other tools exist that focus on manipulating only a specific part of the interaction channel, e.g., changing the speech signal to adjust for emotional cues \cite{rachman2018} or changing gestures recorded via virtual reality tools to produce fake gestures \cite{gurion2018}.
Furthermore, specific environments have been created to present rich interaction contexts to participants while keeping the interaction channel fixed, usually in a turn-by-turn setting. An example is the Minecraft virtual environment in which communication happens between a player and a bot to build objects together \cite{gray2019}.

A previous version of the \slurk \ software \cite{schlangen2018a} has already been used for research, recruiting participants via \amt. For example, \newcite{ilinykh2019a} set up a task in which participants use commands to navigate a network of rooms, represented by changing images; \newcite{attari2019} presented images for participants to discuss; \newcite{galetzka2020}  manipulated the status of messages, controlling whether information is shared or private; and \newcite{chiyahgarcia2020} used a dialog task in which the window layout differs depending on the participant's role in the interaction, including buttons for certain actions. \newcite{haber2020} described their setup for presenting participants with static images without recruitment via a crowdsourcing platform. In this paper, we describe the existing and new features in more detail. In the new software version, we have improved the {\sc api} and extended it to handle audio and video data and implemented a number of new plug-ins and example bots that exemplify the described behaviors.
\section{Architecture and features}
\label{sec:architecture}

\Slurk \ runs as a server that provides the interface to communicate with clients, which can be either human dialog participants or software ``bots''. Interactions happen in ``rooms'' to which the clients log on. A permission system controls what a client can see and do. Room layouts are divided into a \textit{display} area that can be used for providing task-based content such as images, and a \textit{chat} and \textit{input} area that shows the chat history, input field, and the video if desired, see Figure~\ref{fig:dito_image}.

\begin{figure*}
    \centering
    \includegraphics[trim=57 450 100 158,clip,width=0.93\textwidth]{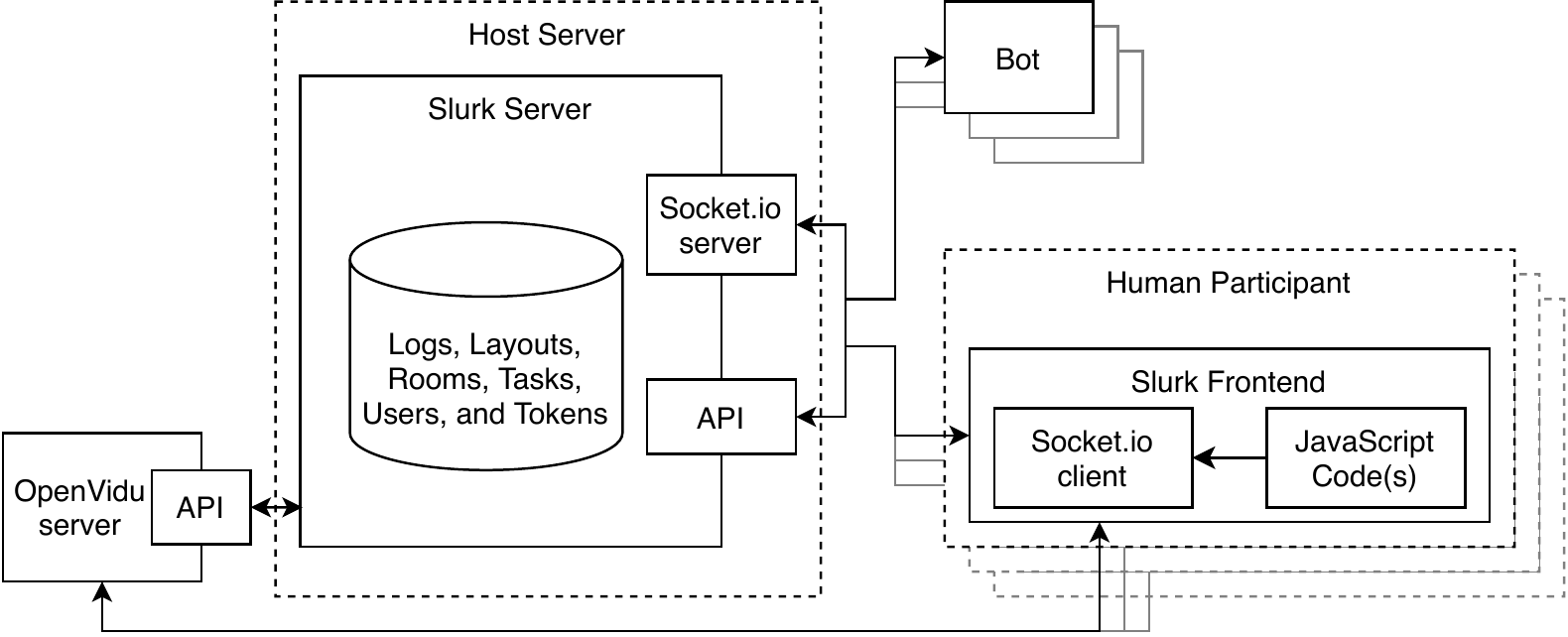}
    \caption{The \slurk \ main architecture. The \slurk \ server is deployed on a host machine. Clients – bots or participant frontends – connect via the \slurk \ {\sc api} and socket.io (all communication between clients happens via the server). Participants log into rooms using a generated token to see the frontend of their assigned room.}
    \label{fig:architecture}
\end{figure*}

\subsection{Core \slurk \ concepts}
The core concepts in \slurk \ are the following:

\begin{itemize}
    \item \textbf{Room}: A room is the space in which users interact with each other, with a bot, or with material presented to them, e.g., an image.
    \item \textbf{User}: A user is a participant in the interaction who has certain permissions that may restrict her access. Permissions may also change during an interaction, e.g., at the end of a task we may want to prevent users from making further contributions. Both human participants and bots are users. A human user can only be in one room at a given time. Bots can participate in several rooms at once.
    \item \textbf{Task}: A task defines what a room looks and behaves like. We can define the number of users to be assigned to a task. Rooms can then be opened based on task information, so that several rooms can have interactions about the same task in parallel.
    \item \textbf{Token}: User permissions are encoded via tokens. Tokens also carry information about which task a user is assigned to.
    \item \textbf{Layout}: The layout defines what the users can see, e.g., what type of chat history, and what the context looks like, e.g., whether an image or buttons are visible. The visible context can change during an interaction.
    \item \textbf{Events}: Both server and clients emit events that bots may react to for defining the logic of an interaction. For example, bots may react to a user entering a room or on a text message that was sent in a certain room.
\end{itemize}

A typical data collection setup involves defining a layout and task and using a script that creates rooms on demand (the Concierge Bot, cf.\ Section~\ref{ssec:concierge}), whenever the necessary number of participants has entered a waiting room. Other bots can then join this room. Human users log in to \slurk \ using a url link that encodes their token and with it the permissions they have for carrying out actions like sending messages or images. We describe some example bots and settings in Section~\ref{sec:example_settings}.

\subsection{Technology}
Figure~\ref{fig:architecture} shows the overall server-client architecture of \slurk. The system is built in Python, on top of flask\footnote{\url{http://flask.pocoo.org}} and flask-socketio.\footnote{\url{https://flask-socketio.readthedocs.io}} The \slurk \ server communicates with a separate video server via https for audio and video communication (cf.\ Section~\ref{ssec:audiovideo}). The video server can be deployed on a different machine and is configured via the \slurk \ \textsc{rest} \textsc{api}, thus making everything configurable in one place.

Bots use the \slurk \ \textsc{api} via socket.io\footnote{For example using Python or any other socket.io client.} to act in a particular setup: They can create rooms on the fly, send users into a room or disconnect them, or they can associate a video session to a room.
Bots use the \textsc{api} also to act as overt or hidden dialog participants. They can react on server events, e.g., when human users send text messages, commands or images, or when they click on an element in the display area. Bots emit events themselves when they message a user or when they modify the display area of one or more users.

Human users are served an \html-page that connects to the \slurk \ server via flask-socketio. A token specifies the permissions they have, e.g., to send private messages or commands or to participate in a video session. Permissions are specified by the experimenter depending on the task.
The frontend that users see is written in \js \ and configures some functionality via plug-ins, e.g., the command syntax and the syntax for private messaging.

\begin{figure*}

    \begin{subfigure}{0.55\textwidth}
    \centering
    \includegraphics[width=\columnwidth,trim=130 497 160 50,clip]{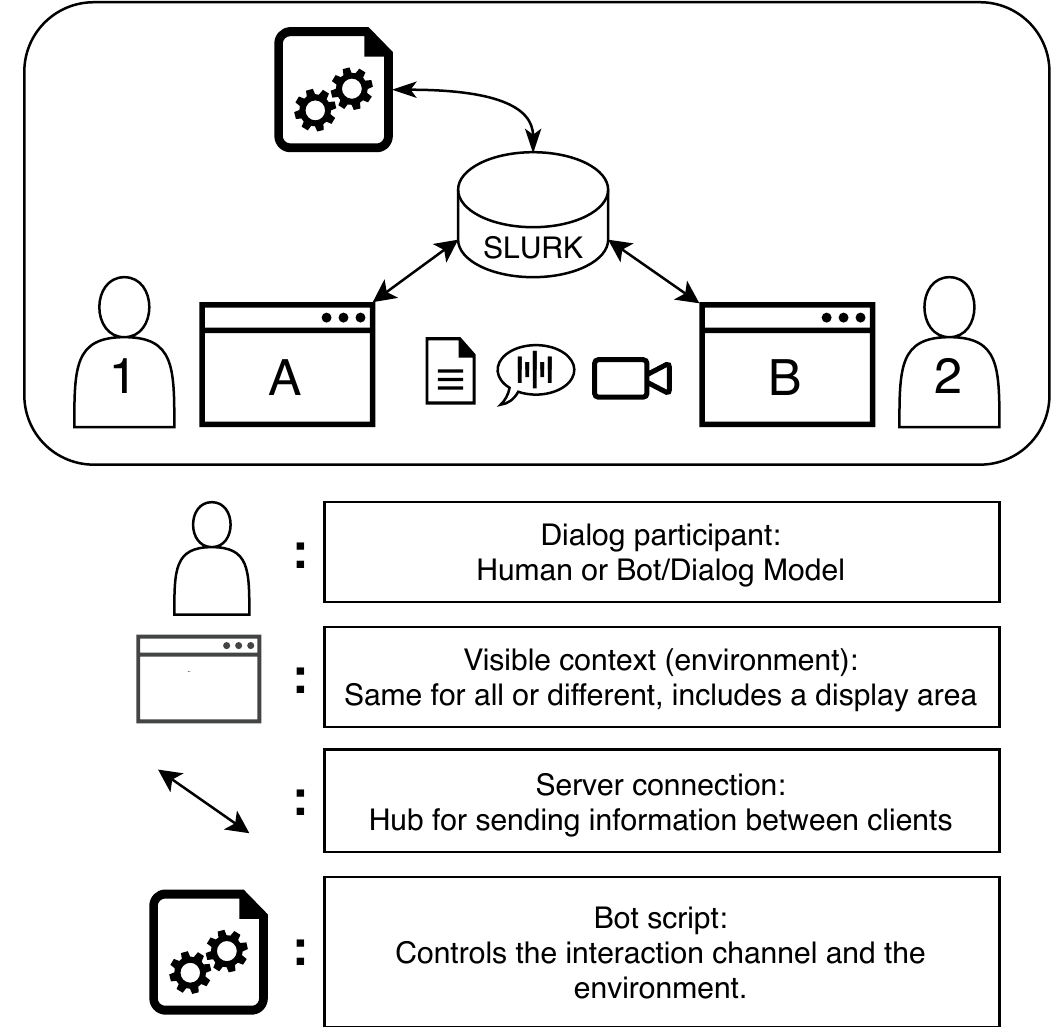}
    \caption{Abstract depiction of possible information flow: Human participants access the server via their browser clients, bot scripts can access the server via the \textsc{api}. In practice, bot scripts and bot participants are implemented as one object.}
    \label{fig:interaction_possibilities}
    \end{subfigure}\hfill
    \begin{subfigure}{0.42\textwidth}
    \includegraphics[width=\columnwidth,trim=144 607 156 57,clip]{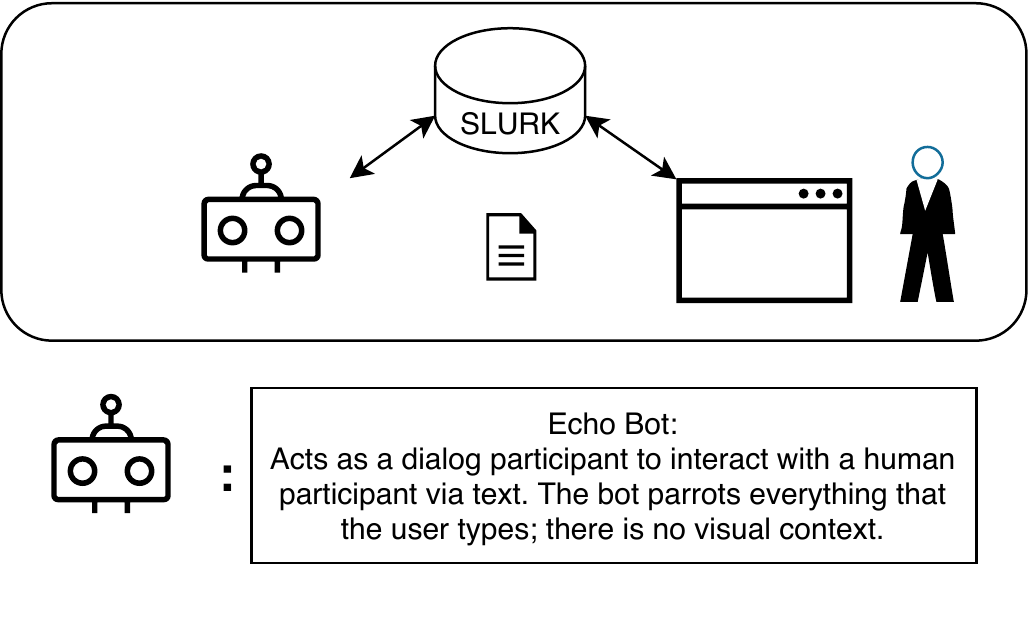}
    \includegraphics[width=\columnwidth,trim=144 597 156 28,clip]{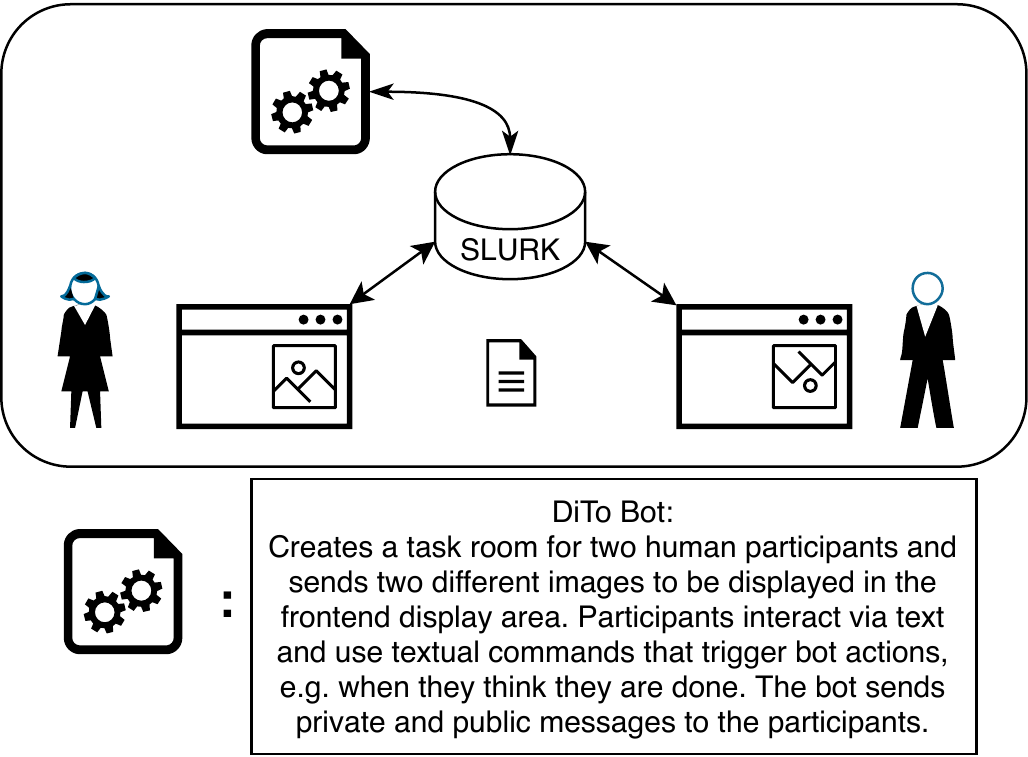}
    \caption{Setups for the Echo Bot with one participant (top) and the DiTo Bot with two participants (bottom).}
    \label{fig:echo-dito}
    \end{subfigure}
    \caption{Schema and two instantiations of interaction setups.}
\end{figure*}

\subsection{Audio \& video interaction}
\label{ssec:audiovideo}

Adding video to remote dialog allows us to bridge the gap between a face-to-face
setting and a pure audio setting with shared visual context. Video analysis of facial gestures in face-to-face conversation has for some time played a role in studying dialog \cite{vanson2008,oertel2013}. Adding the video modality to remote conversation can give us new insights into the role of the facial gesture modality. Note that it is also possible to use \slurk \ in settings where there is only one user to record, e.g., to collect acted gesture performances (video) or spoken descriptions of images or other stimuli (audio only).

For allowing users to interact via audio and video, we have added the possibility to connect to an \openvidu\footnote{\url{https://openvidu.io}} \ server session directly via \slurk. \openvidu \ is an open-source software for streaming live video and audio based on the highly compatible WebRTC framework for streaming multimedia data. It is licensed under Apache License v2 and the free version of the software includes streaming and recording videos, as well as self-hosting, making it possible to keep full control of the data.

Connections to \openvidu \ sessions can be set up directly via the \slurk \  \textsc{api}. For a room to include a video or audio session, it has to be associated with the respective \openvidu \ session. Users in the room must be granted permissions (via their tokens) to publish or subscribe to the \openvidu \ session. In the future, we plan to further explore the quality of this data that we can obtain using crowdsourcing and the privacy implications of this type of data collection.

\subsection{Bots}\label{ssec:bots}

Bots are client programs that can combine different roles in order to control the interaction.
Bots can take the role of a full dialog agent, i.e., we can use them to test any dialog model by letting human users interact with the model. Bots can also control the interaction as hidden agents, e.g., they can interpret commands, modify contributions, or suppress contributions. This gives us the possibility to control an ongoing interaction, e.g., a bot might react to certain words and send a private message to a user in response, or impersonate another user by modifying their turn.

Bots may also implement background logic, i.e., they interpret button presses and other events happening in the display area, react on incoming users, or track time since the last contribution. For example, in the MeetUp!\ corpus \cite{ilinykh2019a}, users navigate a room network by typing commands such as \texttt{/n} for ``go north''. The bot interprets the command and changes the image in the display area accordingly.
Bots can also administrate the main settings, e.g., they can move users between rooms or change their permissions.

\subsection{Configuring the visual context}

A growing body of research is today concerned with building models that can integrate language and visual information, for example to reason about spatial relations of objects in an image \cite{bisk2016a}. \Slurk's display area is configurable to show dialog participants custom images or other visual material. The configuration of various visual aspects is done via a \textsc{json} file from which \html \ is built. The bot script can then  determine (and throughout the interaction change) what each user can see. For example, in the DiTo Bot setting, two participants are trying to find the difference between the images that they see (cf.\ Figure~\ref{fig:echo-dito}). A screenshot of what one participant's browser window looks like can be seen in Figure~\ref{fig:dito_image}.

\subsection{Logging}

\begin{figure}
\centering
  \includegraphics[trim=10 535 1630 100,clip,width=\columnwidth]{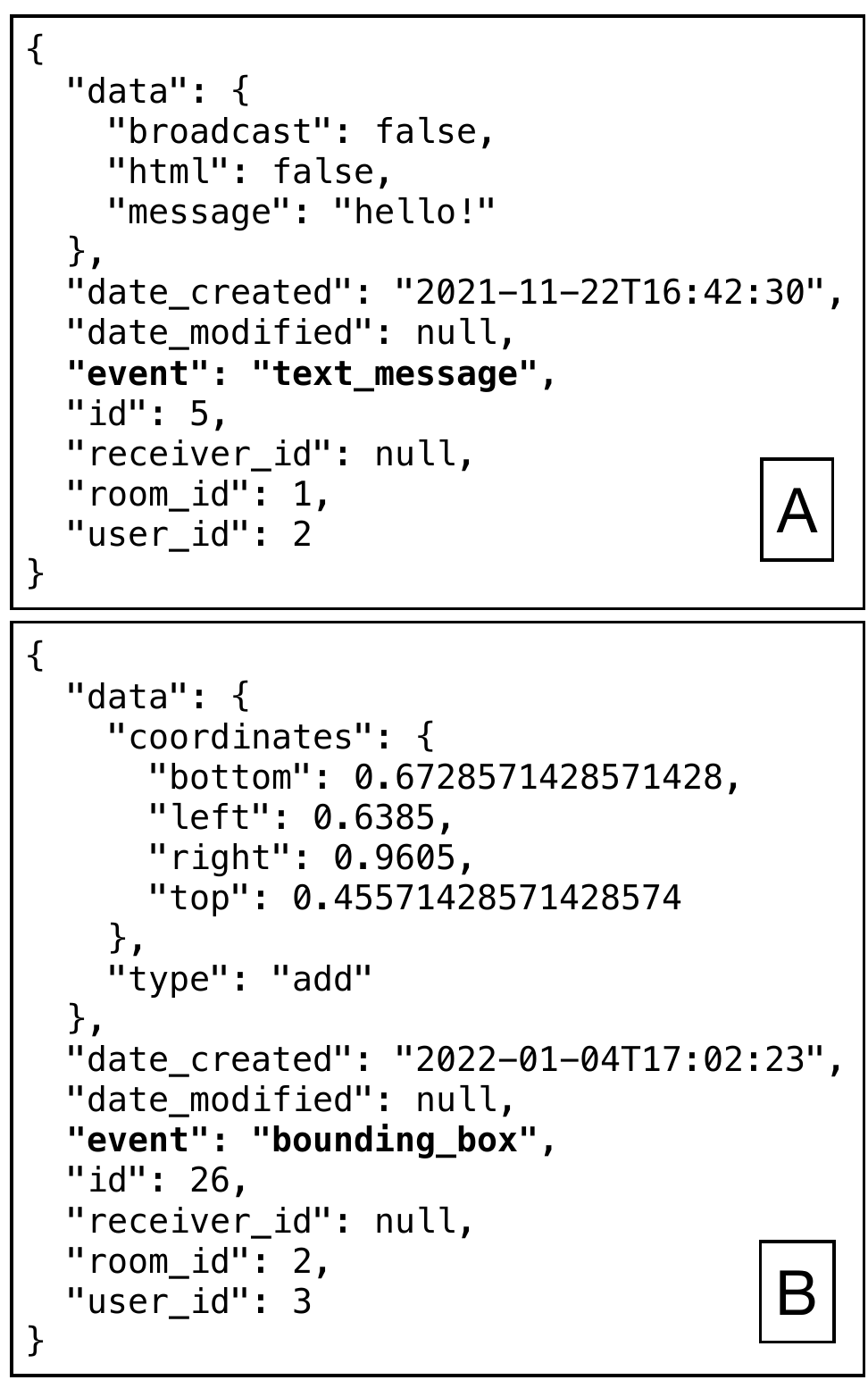}
  \caption{Example log entries for different events: (A) A user has sent a text message. (B) A user has drawn a bounding box in the display area.}
  \label{fig:logging}
\end{figure}

All parts of an interaction are logged on the \slurk \ server in a \textsc{json} format. Figure~\ref{fig:logging} shows example log entries for common events. \textsc{Json} data\footnote{\url{https://www.json.org}} is human-readable and can easily be parsed automatically and converted to other data formats for annotation or analysis. Every log entry contains a timestamp, allowing synchronization with all data sources, including the video and audio recordings. Logging happens continuously, so that logs retrieved at a certain time reflect the interaction up to that point.

\subsection{License, download and development}

\Slurk \ is available at \url{https://github.com/clp-research/slurk} under a BSD-3-Clause License. The repository contains instructions about how to download, install and deploy the server, as well as a tutorial for setting up an example room. A second repository contains example bots at \url{https://github.com/clp-research/slurk-bots}. Both repositories are public for others to contribute to or file issues for support and improvement.
\section{Studying dialog with \slurk}
\label{sec:example_settings}

The \slurk \ tool is intended to allow fine-grained manipulation of interaction behaviors in order to study dialog phenomena such as creating common ground in shared or private settings. A specific dialog experiment will have to define the interaction channel, e.g., how many participants are interacting and whether the interaction is written or spoken, the context that is available to the participants, i.e., what they can see and how they can interact with entities in the context, and the rules of the interaction, e.g., when the interaction is successful. We describe below how the settings for these building blocks can be manipulated in \slurk.

\subsection{Controlling interaction settings}

\begin{figure}
    \begin{center}
    \includegraphics[trim=51 10 25 1, clip, width=\columnwidth]{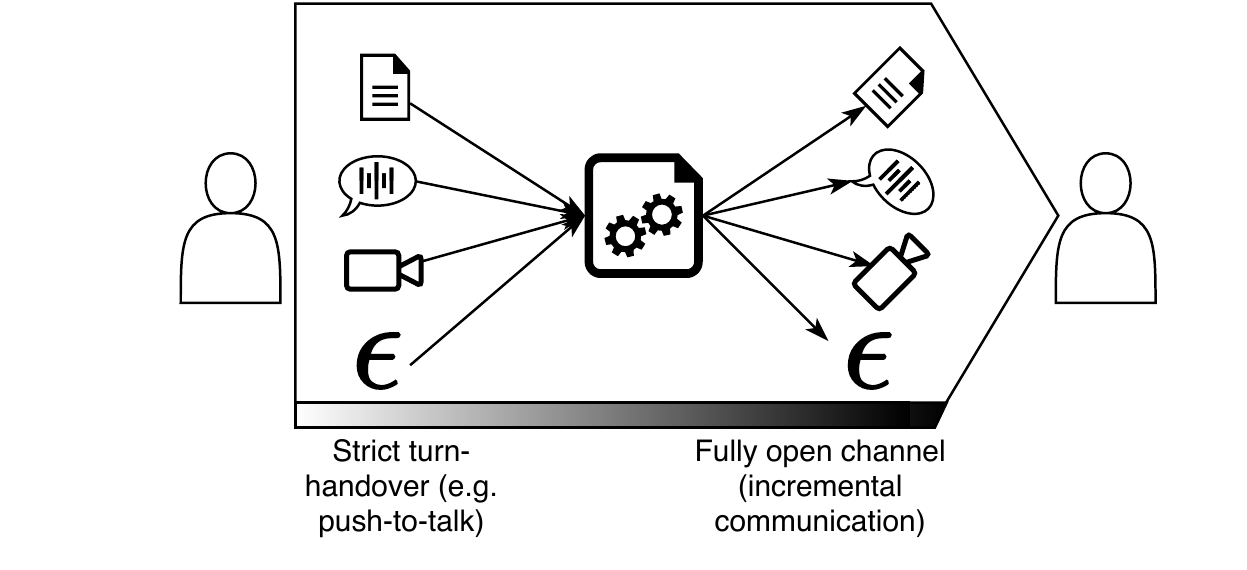}
    \end{center}
    \caption{The interaction channel between users (humans or bots) can be intercepted and modified in a number of ways, including removing and inserting elements (\(\epsilon\)). The bot can either be an overt participant or be invisible to some or all others. Human user clients communicate via the \slurk \ server that emits events to the bot who can then manipulate the channel. Turn-taking regimes can set restrictions on how early a message is sent and when the addressee can answer.}
    \label{fig:interaction_channel}
\end{figure}

Figures~\ref{fig:interaction_possibilities} and \ref{fig:interaction_channel} schematically show how the settings of the chat and display area can be adjusted.
The default mode of interaction is text, shown in the interaction area in the left part of the window. The chat area contains the dialog history as is common for similar chat tools. By default, users can see their own and others' messages, ordered by recency.
In addition, an audio and video connection can be established. Any visual dialog context is shown in the display area in the right part of the browser window (cf.\ Figure~\ref{fig:dito_image}).

\begin{figure}
\begin{subfigure}{\columnwidth}
\centering
\setlength{\fboxsep}{0pt}
\setlength{\fboxrule}{0.1pt}
  \fbox{\includegraphics[width=0.9\columnwidth]{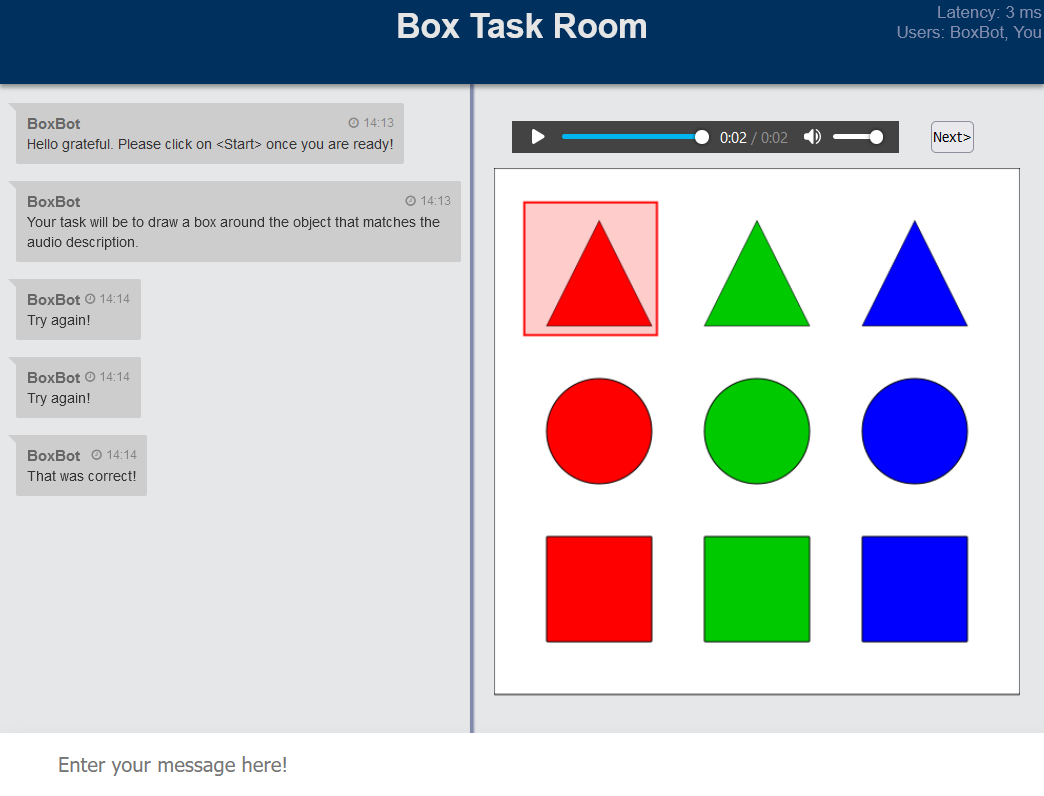}}
  \caption{Display area containing an audio player, a button next to it, an image, and a bounding box that a user has drawn.}
\end{subfigure}
\hfill
\begin{subfigure}{\columnwidth}
\centering
\setlength{\fboxsep}{0pt}
\setlength{\fboxrule}{0.1pt}
  \fbox{\includegraphics[width=0.9\columnwidth]{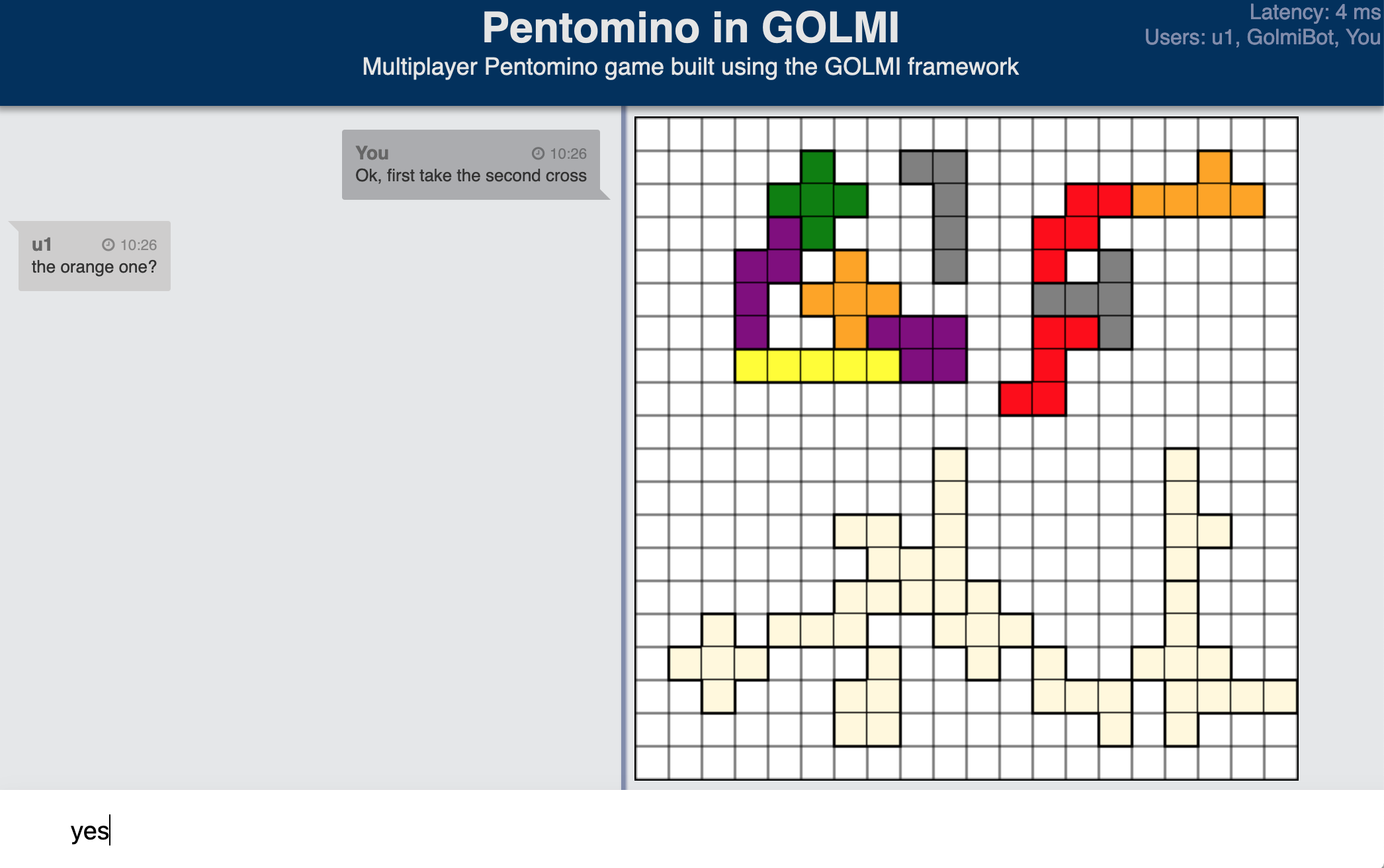}}
  \caption{Display area containing an interactive Pentomino game.}
  \label{fig:pentomino_screenshot}
\end{subfigure}
\caption{Example display areas.}
\label{fig:layout_browser}
\end{figure}

\paragraph{Incrementality:}Two plug-ins are available to show to users how the dialog is evolving:
The {\tt typing-users} plug-in shows who is currently typing, the {\tt live-typing} plug-in sends messages directly as they are being typed, without the typer having to hit the ``send'' button. Plug-ins are specified in the room layout as shown in Figure~\ref{fig:plugin}.
\paragraph{Turn-taking:}Turn-taking can be manipulated along a scale to either follow a strict turn-handover regime in which the addressee cannot talk until they are explicitly given the turn on the one hand, or be fully incremental on the other hand. By default, anyone can type or speak at any time. Bots can temporarily prevent users from typing by changing their permissions based on custom logic. For example, a bot can keep track of turns and enforce a round-robin format in which each participant needs to wait until it is their turn to contribute.\\
Similarly, in spoken interaction, we can control whether the audio channel is always open for everyone to speak or enforce stricter turn-taking by using a ``push-to-talk'' setting in which speakers signal when they are done speaking in order to open the channel for someone else.\footnote{Note that it's possible to mix written and spoken dialog.}
\paragraph{Intercepting the channel:}Communication between the clients happens via the server and it's therefore possible to modify any messages before sending them on to their addressee (cf.\ Figure~\ref{fig:interaction_channel}). Bots can change, insert or delete text messages and pretend to be another user. In the same way, it is possible to intercept the audio and video channels, although we are leaving such tests to future work.
\paragraph{Multimodal context:}The display area serves as dialog context, controlled by \js, and can contain arbitrary \html \ elements. For example, the display area can present images, buttons, or pre-recorded audio and video elements, or embed interactive tools. The position of the live video can be freely adjusted. We also make available sample plug-ins that track mouse movement and let participants draw bounding boxes. Any element in the display area can be modified programmatically by bots during the interaction so that the context can change, e.g., images may become visible or invisible to one or more users. An example is shown in Figures~\ref{fig:layout} and \ref{fig:layout_browser}.
\paragraph{Shared vs.\ private information:}Permissions control a range of possible user behavior, including what each participant can see in the context, i.e., whether all participants see the same context or different contexts. Privately sent text is only visible to the respective addressee. In this way, bots can send administrative messages to single participants, e.g., to encourage them to contribute or remind them of the task rules.

\begin{figure}
\begin{subfigure}{\columnwidth}
\small
\begin{Verbatim}[frame=single]
{
  "title": "Room",
  "scripts": {
    "incoming-text": "display-text",
    "incoming-image": "display-image",
    "submit-message": "send-message",
    "print-history": "plain-history",
    "typing-users": "typing-users"
  }
}
\end{Verbatim}
\caption{A minimal room layout defining what \js \ plug-ins to use for different tasks. In this example, text, images and history are displayed, as well as who is currently typing.}
\label{fig:plugin}
\end{subfigure}
\begin{subfigure}{\columnwidth}
\small
\begin{Verbatim}[frame=single]
{
  "title": "Box Task Room",
  "html": [
  {
    "layout-type": "div",
    "style": "text-align: center;",
    "layout-content": [{
      "layout-type": "audio controls",
      "id": "audio-file",
      "src": "",
      "autoplay": "true",
      "style": "height:30px;"
    }]
  },
  {
    "layout-type": "div",
    "style": "text-align: center;",
    "layout-content": [{
      "layout-type": "image",
      "id": "drawing-area"
    }]
  }],
  "scripts": {
    "incoming-text": "markdown",
    "incoming-image": "display-image",
    "submit-message": "send-message",
    "print-history": "markdown-history",
    "plain": "bounding-boxes"
  }
}
\end{Verbatim}
\caption{A room layout specifying \html \ elements for an audio player and an image. The bot will later insert the respective source files. The image element is labeled with {\tt "id":"drawing-area"} so that the bounding-box plug-in can access the element. The plug-in is specified in the {\tt scripts} block.}
\label{fig:layout}
\end{subfigure}

\caption{Example {\sc json} room layouts.}
\end{figure}

\subsection{Data collection in practice}

\subsubsection{Pairing participants}
\label{ssec:concierge}

When collecting human-human dialog data, participants need to be recruited and paired based on the specific task. The biggest challenge in connection to collecting dialog data from crowdworkers is often the synchronous nature of the task. For most other tasks, crowdworkers perform tasks in their own time, with no dependencies on other workers. This section shows how we address this challenge.

For data collections, we set up a task bot that is responsible for the core of the dialog game users are asked to play. This bot might first present some general instructions to users, repeat the dialog task, or remind a user to engage with a particular part of the context. For example, the DiTo Bot in Figure~\ref{fig:dito_image} (a variant of the bot used by \newcite{attari2019}) presents the task and makes sure both users are ready to start the dialog.

When participants log in to \slurk, they are not immediately sent to a room with their task bot, but instead they see a waiting room and another bot that is in the room with them. This bot, the \textit{Concierge Bot}, monitors incoming users and their tasks and keeps track of the task requirements: Once enough users for a task have entered, the Concierge Bot creates a new task room and sends the users to the new room. The task bot then joins this task room as well. Any bot can also track time, so that participants can receive a small reimbursement even for their waiting time and are sent back to the crowdsourcing platform in case no other participants should appear within a given timeframe, e.g., five minutes (cf.\ Section~\ref{ssec:amt}).

For the DiTo task, two participants are required, so the Concierge Bot forwards incoming users to a newly created task room once this number is reached. A third and fourth participant would trigger the Concierge Bot to create a second task room, so that the same task can be carried out in parallel in different rooms. Figure~\ref{fig:dito_image} shows the task room setup. Figure~\ref{fig:echo-dito} schematically shows the task room setup that configures the interaction channel and display area in which users have separate visual contexts and a bot is present to administrate the interaction by tracking time and counting contributions.

\subsubsection{Connecting with crowdsourcing platforms}
\label{ssec:amt}

Another requirement for integrating a \slurk \ data collection on a crowdworking platform such as Amazon Mechanical Turk is that workers need to leave the platform for performing the task and after completing it need to return to the platform, so that their data can be associated to the dialog logs in order to evaluate and reimburse them correctly.

When directing participants from the crowdworking or experiment platform to their \slurk \ room, it's possible to include all necessary information into the url they need to follow. In particular, the url contains their individual access token and we can automatically assign them a name that keeps them anonymous. This has the advantage of reducing the steps they would otherwise need to follow (choosing an access name and copy/pasting the token).

In order to connect their dialog data (present in the \slurk\ server) to their crowdworking profile (present in the provider's system), a mechanism can be included in a bot that provides an individual code to each participant once they have completed their task. The code needs to be entered on the crowdworking platform once they have returned and links their profile with the correct dialog logs.

\subsubsection{Collecting video data via crowdsourcing}

Collecting dialog data remotely involves pairing people that usually do not know each other. While random pairing has been done for both text and audio, e.g., for the Switchboard corpus \cite{godfrey1992}, the video channel adds another dimension of privacy concerns that needs to be accounted for.

Video data has been collected via \amt \ before, e.g., by \newcite{sigurdsson2016}, but it is a different matter whether a participant shares video with researchers that can vouch for data security or whether to share live video with another participant. Special care also needs to be exacted in deploying the video server to safeguard it from unauthorized access.

The \openvidu \ platform allows us to record video on our own servers and adhere to local data protection laws. We explicitly want to stress that when using the audio and video features, participants need to be informed that their data is recorded in these modalities and that their explicit consent is needed.

\section{Conclusion}

We have described \slurk, a lightweight interaction server to experiment with dialog. The infrastructure can be used to collect dialog data or test dialog models by letting human participants interact with them. \slurk \ allows a range of manipulations to the interaction channel as well as user-defined dialog context, such as images or interactive elements.
Among the planned work for the future are a bot that allows collaborative
manipulation of a Pentomino game board \cite{zarriess2016a}, cf.\ Figure~\ref{fig:pentomino_screenshot}; an integration for a
text messenger such as Telegram; and experimenting further with the audio and video channel.
We invite the community to use the tool for their purposes and are open for
suggestions for further features.
We hope that the tool can contribute to making it easier to collect better
dialog data that in turn lead to better models.

\section{Acknowledgements}
This work was partially funded by DFG project 423217434 (``recolage'').

\section{Bibliographical References}\label{reference}

\bibliographystyle{lrec2022-bib}
\bibliography{slurkbib}

\end{document}